\documentclass[x11names]{elsarticle}
\usepackage{tikz}
\usetikzlibrary{shapes,arrows,chains}
\usepackage[utf8]{inputenc}
\usepackage{lineno,hyperref}
\usepackage{booktabs}
\usepackage{adjustbox}
\usepackage{multirow}
\usepackage{lineno}
\modulolinenumbers[5]
\usepackage{placeins}
\usepackage{graphicx}
\usepackage[font=small,labelfont=bf]{caption}
\usepackage{subcaption}
\usetikzlibrary{shapes, shadows, arrows, positioning}
\usepackage{algorithm}
\usepackage{algpseudocode}
\usepackage{hyperref} 

\begin{document}

\begin{frontmatter}

\title{A $k$ nearest neighbours classifiers ensemble based on extended neighbourhood rule and features subsets}

\author[add1]{Amjad Ali}
\author[add1]{Muhammad Hamraz}
\author[add1]{Naz Gul}
\author[add1]{Dost Muhammad Khan}
\author[add1]{Zardad Khan \corref{mycorrespondingauthor}}
\cortext[mycorrespondingauthor]{Corresponding author}
\ead{zardadkhan@awkum.edu.pk}
\author[add2]{Saeed Aldahmani}

\address[add1]{Department of Statistics, Abdul Wali Khan University Mardan, Pakistan}
\address[add2]{Department of Analytics in the Digital Era, United Arab Emirates University, UAE}

\begin{abstract}
$k$NN based ensemble methods minimise the effect of outliers by identifying a set of data points in the given feature space that are nearest to an unseen observation in order to predict its response by using majority voting. The ordinary ensembles based on $k$NN find out the $k$ nearest observations in a region (bounded by a sphere) based on a predefined value of $k$. This scenario, however, might not work in situations when the test observation follows the pattern of the closest data points with the same class that lie on a certain path not contained in the given sphere. This paper proposes a $k$ nearest neighbour ensemble where the neighbours are determined in $k$ steps. Starting from the first nearest observation of the test point, the algorithm identifies a single observation that is closest to the observation at the previous step. At each base learner in the ensemble, this search is extended to $k$ steps on a random bootstrap sample with a random subset of features selected from the feature space. The final predicted class of the test point is determined by using a majority vote in the predicted classes given by all base models. This new ensemble method is applied on 17 benchmark datasets and compared with other classical methods, including $k$NN based models, in terms of classification accuracy, kappa and Brier score as performance metrics. Boxplots are also utilised to illustrate the difference in the results given by the proposed and other state-of-the-art methods. The proposed method outperformed the rest of the classical methods in the majority of cases. The paper gives a detailed simulation study for further assessment.
\end{abstract}

\begin{keyword}
\texttt{Features subset, Nearest Neighbours Rule, $k$NN Ensemble, Classification.}
\end{keyword}

\end{frontmatter}

\section{Introduction}
\label{Introduction}
Classification is a supervised learning problem dealing with distributing samples into different classes based on various features. There are several machine learning procedures used for classification, the most popular of which is the nearest neighbour (NN) method \cite{cover1967nearest}. It classifies an unseen observation based on its neighbourhood in the feature space. Nearest neighbour is an efficient method, but has the problem of over-fitting. To overcome this problem, the $k$ nearest neighbour ($k$NN) classifier was proposed which extends the nearest neighbourhood to more than one training observation \cite{cunningham2021k, altman1992introduction, hastie2009friedman}, using the majority vote to classify an unseen instance. This method is simple, easy to understand and provides efficient results when the dataset is sufficiently large \cite{abbasifard2014survey, amal2011survey, kulkarni2013introspection}. Despite being computationally simple, the $k$NN model gives optimal results in many cases and even trounces other complex and composite classifiers. However, $k$NN procedures suffer from many data related issues, such as noise and contrived features in the dataset.

$k$NN ensemble-based learners, in conjunction with randomization procedures, have demonstrated efficient prediction performance. Randomization is usually incorporated by taking random bootstrap samples from training observations and/or random subsets from the total number of features to construct the base $k$NN models. This decreases the chance of repeating the same error and makes the base models more flexible and diverse \cite{bay1999nearest, kaneko2000combining, domeniconi2004nearest, garcia2009boosting}. Several $k$NN based ensembles have been proposed in the literature, e.g. random $k$-NN \cite{li2014random}, ensemble of random subspace $k$NN \cite{rashid2021random}, ensemble of subset of $k$NN \cite{gul2018ensemble}, bootstrap aggregated $k$-NN \cite{steele2009exact}, weighted heterogeneous distance Metric \cite{ZHANG201913}, etc. These methods use majority voting based on the class labels of sample points in the neighbourhood of a given test observation determined by each primary learner. Final prediction is calculated by using a second round of majority voting based on the results given by all the base $k$NN models. However, this type of prediction, based on the nearest neighbourhood rule, might be effected when an unseen observation follows a pattern that goes beyond the sphere containing the nearest observations. Therefore, in such situations, it is desired to devise a new neighbourhood rule which allows for identifying patterns on the far side of the conventional sphere.

Following the above notion, this work proposes a new extended neighbourhood rule (ExNRule) for $k$NN ensemble, where each base $k$NN model is constructed on a random bootstrap sample drawn from training observations in conjunction with a randomly selected subset of features. The ExNRule searches for similar patterns on extended paths i.e. it determines the nearest point $X^1_{1\times{p'}}$ to the test point $X^0_{1\times{p'}}$, then it finds the nearest point $X^2_{1\times{p'}}$ to the previously identified point $X^1_{1\times{p'}}$, and so on. This process continues until the desired $k$ observations are identified whose class labels are used to predict the target class of the test point $X^0_{1\times{p'}}$ using majority voting. Final estimated class of $X^0_{1\times{p}}$ is obtained by majority voting based on the results given by base models. For assessing the performance of the proposed ensemble, 17 benchmark datasets are used, and the resulting performance metrics of accuracy, Kappa and Brier score (BS) are compared with those of $k$NN, weighted $k$ nearest neighbours classifier (W$k$NN), random $k$ nearest neighbour (R$k$NN), random forest (RF), optimal trees ensemble (OTE) and support vector machine (SVM). For further illustration, boxplots have also been obtained to demonstrate the difference in the performance of the proposed ExNRule and other classical procedures.

The remainder of this paper is organized as follows. Related work is summarized in Section \ref{literature}. Section \ref{method} presents a discussion of the proposed method and the associated mathematical descriptions and algorithm. Experiments and results are given in Section \ref{Experiments}. Finally, a conclusion of the analyses conducted in this paper are given in Section \ref{conclusion}.

\section{Related Work}
\label{literature}
Extensive research has been carried out to improve the performance of classical $k$NN classifier. Due to the fact that the classical $k$NN procedure gives equal weights to all $k$ neighbours of a new observation, Bailey et al. \cite{bailey1978note} suggested a weighted $k$NN procedure to improve the standard $k$NN method. In this case, weights are assigned to the neighbours based on their distances from a query point. This procedure is global in that it uses all training instances; therefore, it takes more execution time. Alpaydin \cite{alpaydin1997voting}, Angiulli \cite{angiulli2005fast} and Chidananda et al. \cite{gowda1979condensed} proposed the condensed nearest neighbour (CNN) to reduce data size and to boost up the running time by removing identical samples that do not provide extra information. However, CNN depends on the data order, which may lead to ignoring observations lying on the boundary (extreme observations). Gyeoffrey et al. \cite{gates1972reduced} proposed a similar procedure known as the reduced nearest neighbour (RNN) algorithm by removing samples from training data which do not affect classification performance. In this procedure, templates are removed and training data are reduced. However, like CNN, RNN is also computationally complex.

Another model based $k$NN procedure is proposed in Guo et al. \cite{guo2003knn} to improve the prediction performance and reduce the size of training data. However, this procedure fails in the case of class imbalance and when marginal data out of the identified region is not taken into account. Authors in \cite{yong2009improved} proposed a clustered $k$NN approach to overcome the problem of uneven distribution of training observations, which is a more robust method in nature as compared to the other procedures suffering from class imbalance. However, this method has several deficiencies, the most important of which lies in the difficulty of finding the selection threshold used for distances among a cluster. Moreover, the criteria used to determine $k$ values for different clusters are also unknown. In \cite{parvin2008mknn}, a modified $k$NN algorithm is suggested to use the weights and validity of the training data observations to classify a test observation. The author in \cite{sproull1991refinements} divided the total training dataset in half to develop the $k$-d tree nearest neighbour and used it for the formation of multi-dimensional observations. This method is fast, simple, and easy to understand, and it produces a perfectly balanced tree. However, the $k$-d tree nearest neighbour needs intensive search, is computationally complex and misses the data pattern because it blindly slices training sample points into half. A hybrid method is therefore proposed in \cite{zhang2006svm} based on SVM and $k$NN, which deals naturally with multi-class problems and gives better performance. Further developments of the $k$NN based methods can be found in \cite{chen2016effectively, rohban2012supervised, wu2002improved, altinccay2007ensembling, tahir2007simultaneous}.

In addition to the above literature, there are several ensemble procedures based on $k$NN models that aim to further improve the performance of the base $k$NN and its modified versions. Bao et al. \cite{bao2004combining} have used different metrics for distance calculation, such as perturbations parameters, to introduce diversity in the ensemble. The authors in \cite{ishii2005combining} have suggested to combine different base $k$NN learners using various distance function weights acquired by a genetic algorithm. Ho \cite{ho1998nearest} proposed a component $k$NN algorithm using various random subspaces, where each base $k$NN model is constructed on a subset of features randomly taken from the total feature space. Bootstrap sampling and attribute filtering with random configuration distance functions are used for ensemble $k$NN models in  Zhou and Yu \cite{zhou2005ensembling}, where simultaneous perturbations are applied on attribute space, learning parameters and training data. A genetic algorithm is used by Altinçay \cite{altinccay2007ensembling} to develop an evidentiary $k$NN ensemble procedure presenting multimodal perturbation. In this method, each chromosome statutes a complete ensemble. An efficient perturbation multimodal procedure based on particle swarm optimization is proposed in Nanni and Lumini \cite{nanni2009particle}, where a random subspace method is employed to perturb the feature space and perturbation multimodal procedure.

One of the top ranked ensemble procedures is bootstrap aggregation (bagging) \cite{breiman1996bagging}, which attempts to find the exact bootstrap expectation of the model \cite{caprile2004exact, zhou2005adapt, zhou2005ensembling}. This procedure is the building block for several state-of-the-art ensembles. In this method, hundreds of base learners are built each on a random bootstrap sample drawn from the training observations. The class label for a test point is estimated by majority voting based on the results given by all base models \cite{breiman1996bagging}. In \cite{steele2009exact}, the author modified the exact bagging idea to bootstrap sub-sampling with and without replacement schemes. Several ensemble procedures are constructed that use bagging with a random subset of features for fitting base $k$NN learners \cite{li2014random, gul2018ensemble, gu2018random}. Many authors proposed several techniques to optimize the $k$ value in the base $k$NN classifiers for ensemble methods \cite{grabowski2002voting, zhang2017efficient}. Boosting $k$NN, which is proposed in \cite{garcia2009boosting}, uses two strategies; first, it selects a subspace from the full space, and, second, the inputs are transformed using non-linear projections of the feature space. Further improvements on the boosting methods can be seen in \cite{freund1996experiments, o2000featureboost, zhang2019novel, amores2006boosting, gallego2018clustering}.

Furthermore, there are several ensembles based on $k$NN using different approaches for accurately predicting test data. The optimal $k$NN ensemble given in \cite{ali2020k} fits a step-wise regression model on $k$ nearest observations in each base $k$NN for a test point. Tang and Haibo \cite{tang2015enn} have proposed a method which estimates test data class labels according to the maximum gain of intra-class coherence. Another method similar to the proposed method in this paper is the extended nearest neighbour (ENN) that predicts the target class of a test observation in a two-way communication manner. ENN does not rely only on the observations in the neighbourhood of the new point, but also takes into consideration the spheres containing the new observation as one of their nearest neighbour \cite{tang2015enn}.

The proposed algorithm in this paper is a $k$ nearest neighbour based ensemble where the $k$ neighbours are determined in a stepping manner. Starting from the first nearest observation of the test point, the algorithm identifies a single observation that is closest to the instance identified at the previous step. In all primary learners in the ensemble, this search is extended to $k$ steps on bootstrap samples each with a random subset of the total feature space. Selecting a feature subset for each base model is done to avoid over-fitting and add diversity to the ensemble in addition to that added by bootstrapping.  The final predicted class of the test point is determined by using majority voting based on the predicted classes given by all the primary learners. The proposed procedure improves the estimation in the following ways:
\begin{enumerate}
	\item Each base $k$NN is constructed on a bootstrap sample drawn from the training samples with a random subset of features taken from the total feature space, making the method diverse and preventing the problem of repeating the same errors.
	\item $k$ nearest observations are selected in a step-wise manner to find the true pattern of the test point.
\end{enumerate}

\section{The extended neighbourhood rule (ExNRule) for $k$NN ensemble}
\label{method}
Consider $\mathcal{L} = (X, Y)_{n\times (p+1)}$ to be a training set of data, where $X_{n\times p}$ is a matrix with $p$ features and $n$ sample points and $Y$ is a binary categorical response. Let $X_{1\times p}^{0}$ be a test/unseen sample point with $p$ values and it is needed to predict the output class i.e. $\hat{Y}$ for $X_{1\times p}^{0}$. Suppose $B$ bootstrap samples are drawn from the training data $\mathcal{L} = (X, Y)_{n\times (p+1)}$, each with a random subset of $p'\le p$ features, i.e., $S_{n\times (p'+1)}^b$, where, $b = 1, 2, 3, \ldots, B$ and $X_{1\times p'}^{0}$ is a subset of $p' \le p$ corresponding values from $X_{1\times p}^{0}$. Find the the nearest observation $X_{1\times p'}^{i}$ to $X_{1\times p'}^{i-1}$, where $i=1,2,3, \ldots, k$, by using a distance formula in all $B$ bootstrap samples. Note the corresponding response values of the selected observations, i.e., $y^1, y^2, y^3, \ldots, y^k$ of $X_{1\times p'}^{1}, X_{1\times p'}^{2}, X_{1\times p'}^{3}, \ldots, X_{1\times p'}^{k}$. To get the estimated class of $X_{1\times p}^{0}$, majority voting will be used, i.e., $\hat{Y}^b $ is the majority vote of $y^1 , y^2, y^3, \ldots,y^k$, where, $b=1, 2, 3, \ldots, B$. The final predicted class of the test point $X_{1\times p}^{0}$ is a second round majority vote of $\hat{Y}^1 , \hat{Y}^2, \hat{Y}^3, \ldots, \hat{Y}^B$ i.e. $ \hat{Y} $.

\subsection{Mathematical Description}

The distance formula to be used in $S_{n\times (p'+1)}^b$, where, $b = 1, 2, 3, \ldots, B$, to compute a set of closest observations in a sequence is give below
\begin{equation}
\delta_b(X_{1\times p'}^{i-1}, X_{1\times p'}^i)_{min}  = [\sum_{j=1}^{p'}|x_j^{i-1} - x_j^i|^{q}]^{1/q},   i=1, 2, \ldots, k.
\label{eq1}
\end{equation}

Is this a standard notation? There is no minimization in the expression on the right of the equal sign.

In each base model, the distance formula given in Equation \ref{eq1} is used to determine the sequence of distances as

$$\delta_b(X_{1\times p'}^{0}, X_{1\times p'}^1)_{min}, \delta_b(X_{1\times p'}^{1}, X_{1\times p'}^2)_{min}, \delta_b(X_{1\times p'}^{2}, X_{1\times p'}^3)_{min},$$
$$\ldots, \delta_b(X_{1\times p'}^{k-1}, X_{1\times p'}^k)_{min}.$$

This sequence suggests that, $X_{1\times p'}^{i}$ is the nearest observation to $X_{1\times p'}^{i-1}$, where, $i = 1 , 2, 3, \ldots, k$. The corresponding response values of $X_{1\times p'}^{1},$ $X_{1\times p'}^{2},$ $X_{1\times p'}^{3},$ $\ldots,$ $X_{1\times p'}^{k}$ are $y^1 , y^2, y^3, \ldots,y^k$, respectively, and the predicted class of test point $X_{1\times p}^{0}$ for the $b^{th}$ base model is $\hat{Y}^b = $ majority vote of $(y^1,$ $y^2,$ $y^3,$ $\ldots,$ $y^k)$, where, $b=1, 2, 3, \ldots, B$. The final predicted class of the test observation $X_{1\times p}^{0}$ is $\hat{Y} =  $ majority vote of $(\hat{Y}^1 , \hat{Y}^2, \hat{Y}^3, \ldots, \hat{Y}^B)$.

\begin{figure}[h]
	\centering
	\includegraphics[width=1\textwidth, height=0.35\textheight]{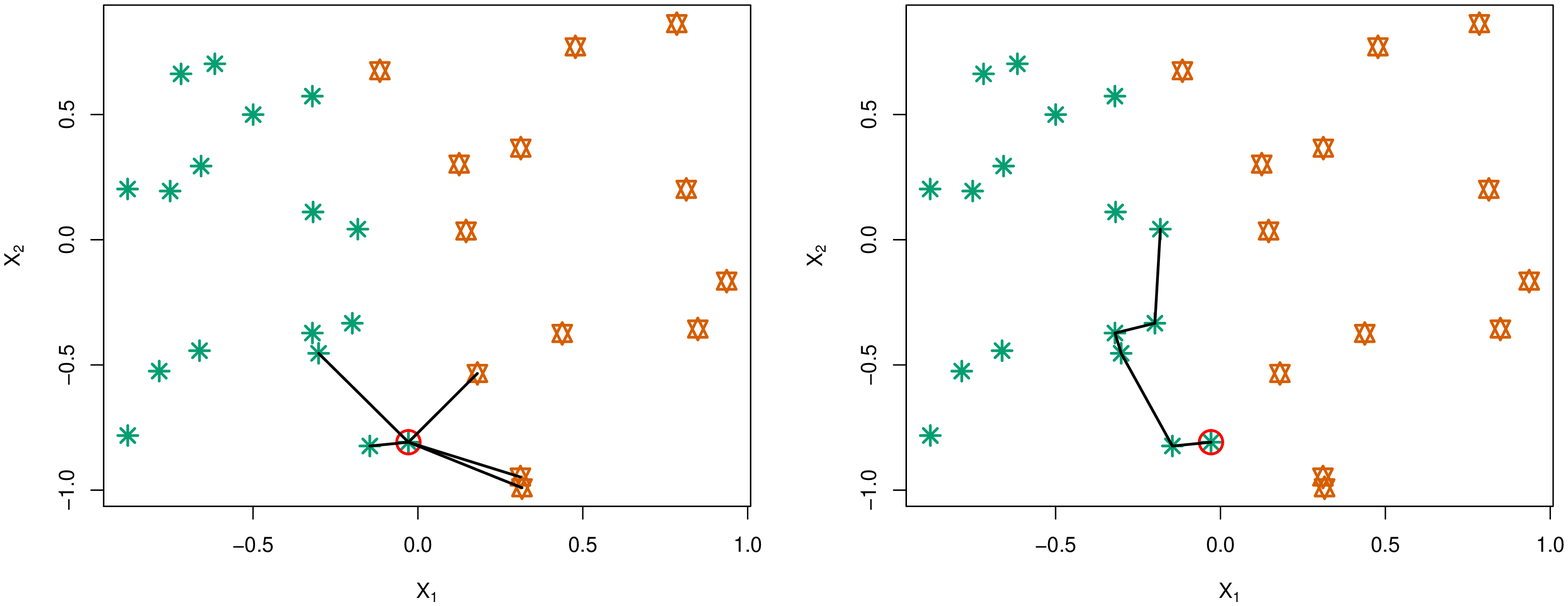}
	\caption{Comparison of the proposed method with usual $k$NN}
	\label{comp}
\end{figure}

A graphical illustration of the proposed ExNRule is given in Figure \ref{comp} against the standard $k$NN model. The figure shows a binary class problem highlighted in grey and green colours. Consider the test observation with true class as green (shown in red circle), whose class label estimate is desired by the models. As can be seen in the figure, the ExNRule has identified observations (shown in green) having the same class as the test point (shown in red circle). The standard $k$NN rule is misleading in this example as the class membership probability of the test point is $0.4$ for the green class and $0.6$ for the grey class, classifying the test point to the grey class. On the other hand, in the case of the ExNRule, the class membership probability estimate of the test point is $1$ for the green class and $0$ for the grey class, classifying the test point to the green class.

\begin{algorithm}
		\begin{algorithmic}[1]
			\caption{Psudue code of the proposed method}
			\State $X_{n \times p} \gets $ Data matrix with $p$ variables and $n$ observations.
			\State $y_n \gets $ Response vector of $n$ values.
			\State $X^0_{1 \times p} \gets $ A test point with $p$ values.
			\State $B \gets $ Total number of random bootstrap samples drawn from training observations.
			\State $k \gets $ Total number of nearest steps on extended paths.
			\State $p \gets $ Total number of variables included in the data.
			\State $p' \gets $ Size of subset of features selected for base models; where $p'\le{p}$.
		
			\For {$b\gets 1:B$}
			
			\State $ S_{n \times p'} \gets $ Bootstrap with $p'\le{p}$ features from $X_{n \times p}$
            \State $ X^0_{1 \times p'} \gets $ Subset of $p'\le{p}$ values from test point $X^0_{1 \times p}$

			\For {$i\gets 1:k$}
			\State $X^i_{1 \times p'} \gets $ Closest training observation to $X^{i-1}_{1 \times p'}$ in $S_{n-(i-1) \times p'}$
			\State $y^i \gets$ The corresponding response value
			\EndFor
			
			\State $\hat{Y}^b = $ majority vote of $(y^1 , y^2, y^3, \ldots,y^k)$
			
			\EndFor
			\State $\hat{Y} =  $ majority vote of $(\hat{Y}^1 , \hat{Y}^2, \hat{Y}^3, \ldots, \hat{Y}^B)$
			\label{algo}
		\end{algorithmic}
\end{algorithm}

\begin{figure}[H]
	\centering
	\fbox{
		\colorlet{lcfree}{Green3}
		\colorlet{lcnorm}{Blue3}
		\colorlet{lccong}{Red3}
		\pgfdeclarelayer{marx}
		\pgfsetlayers{main,marx}
		\providecommand{\cmark}[2][2]{%
			\begin{pgfonlayer}{marx}
				\node [nmark] at (c#2#1) {#2};
			\end{pgfonlayer}{marx}
		}
		\providecommand{\cmark}[2][]{\relax}
		\begin{tikzpicture}[%
			>=triangle 60,              
			start chain=going below,    
			node distance=5.5mm and 50mm, 
			every join/.style={norm},   
			]
			
			\tikzset{
				base/.style={draw, on chain, on grid, align=center, minimum height=5ex},
				proc/.style={base, rectangle, text width=12em},
				test/.style={base, diamond, aspect=2, text width=9em},
				term/.style={proc, rounded corners},
				coord/.style={coordinate, on chain, on grid, node distance=6mm and 30mm},
				nmark/.style={draw, cyan, circle, font={\sffamily\bfseries}},
				norm/.style={->, draw, lcnorm},
				free/.style={->, draw, lcfree},
				cong/.style={->, draw, lccong},
				it/.style={font={\small\itshape}}
			}
			
			\node [proc, fill=Blue3!10, join] (t) {$L=(X, Y)_{n\times{p+1}}$ be training data and  $X_{1\times p}^0$ be a test point};
			\node [test, join, right=of t, fill=lccong!10] (t0) {\textbf{For} $b=1:B$};
			\node [proc,  fill=lcfree!10] (t1) {Take a bootstrap sample $S_{n \times p'}$ with $p'\le p$ features and $X_{1 \times p'}^0$ be a subset of values from $X_{1 \times p}^0$};
			\node [test, join, left=of t1,  fill=lccong!10] (t2) {\textbf{For} $i=1:k$};
			\node [proc, fill=lcfree!10] (t3) {Find the nearest observation $X_{1\times p'}^i$ to  $X_{1\times p'}^{i-1}$ in $S_{n \times p'}$};
			\node [proc, fill=lcfree!10] (t4) {Note the corresponding response ($y^i$) of $X_{1\times p'}^i$};
			\node [proc, right=of t4, fill=lcfree!10] (t5) {$\hat{Y}^b =$ majority vote of $(y^1, y^2, y^3, \ldots, y^k )$};
			\node [proc, fill=Blue3!10] (t6) {$\hat{Y} =$ majority vote of $(\hat{Y}^1, \hat{Y}^2, \hat{Y}^3, \ldots, \hat{Y}^B)$};
			\node [coord, left=of t4] (c1)  {}; 
			\node [coord, right=of t5] (c2)  {};
			\path (t0.south) to node [near start, xshift=1em] {} (t1);
			\draw [->,lcnorm] (t0.south) -- (t1);
			\path (t2.south) to node [near start, xshift=1em] {} (t3);
			\draw [->,lcnorm] (t2.south) -- (t3);
			\path (t3.south) to node [near start, xshift=1em] {} (t4);
			\draw [->,lcnorm] (t3.south) -- (t4);
			\path (t4.east) to node [near start, xshift=1em] {} (t5);
			\draw [->,lcnorm] (t4.east) -- (t5);
			\path (t5.south) to node [near start, xshift=1em] {} (t6);
			\draw [->,lcnorm] (t5.south) -- (t6);
			\path (t4.west) to node [yshift=-1em] {} (c1);
			\draw [->,lcnorm] (t4.west) -- (c1) |- (t2);
			\path (t5.east) to node [yshift=1em] {} (c2);
			\draw [->,lcnorm] (t5.east) -- (c2) |- (t0);
			-------------------------------------
	\end{tikzpicture}}
	\caption{Flowchart of the proposed method}
	\label{flowchart}
\end{figure}
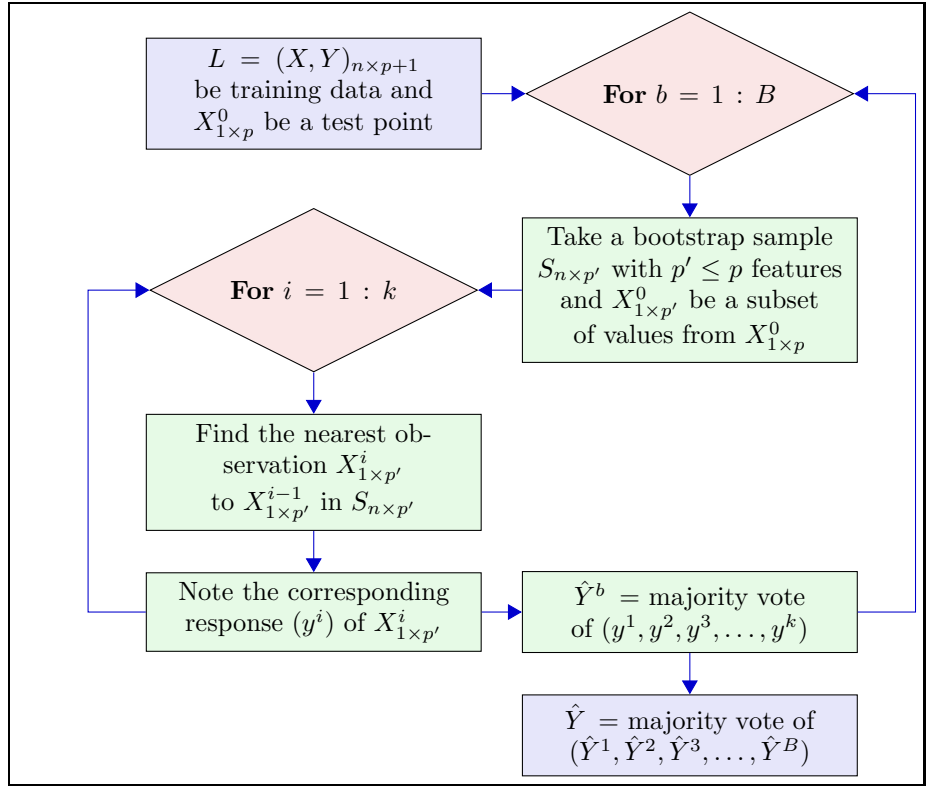

\section{Experiments and Results}
\label{Experiments}
The section presents the conducted experiments and their results for assessing the performance of the proposed ExNRule and  other state-of-the-art methods.

\subsection{Benchmark Datasets}
\label{data5}
A total of 17 benchmark datasets are considered for the analysis of the proposed method and other well known procedures. These datasets are openly available on different repositories, such as openML, UCI, etc. Table \ref{datasets} provides the detailed description of the characteristics of these datasets, i.e., the names of datasets, number of variables, number of instances, class distribution (i.e. 0, 1) and the corresponding sources. The number of features ranges from 7 to 86, while that of observations is from 36 to 583.

\begin{table}[H]
	\centering
	\setlength{\tabcolsep}{3.5pt}
	\fontsize{9.5}{9.5}\selectfont
	\renewcommand{\arraystretch}{1.5}
	
	\caption{A short description of the datasets used in this research.}
	\begin{tabular}{cccccc}
		\toprule
		Data ID &  Data &  $p$ & $n$ & Class distribution & Source \\ \midrule
		$D_1$ & KC1B & 86 & 145 & (85, 60) & \url{https://www.openml.org/d/1066} \\
		$D_2$  &  TSVM & 80 & 156 & (54, 102) & \url{https://www.openml.org/d/41976} \\
		$D_3$  & JEdit &  8 & 369 & (165, 204) & \url{https://www.openml.org/d/1048} \\
		$D_4$  & Cleve & 13 & 303 &  (165, 138) & \url{https://www.openml.org/d/40710} \\
		$D_5$ & Wisc & 32 & 194 & (104, 90) & \url{https://www.openml.org/d/753} \\
		$D_6$ & AR5 & 29 & 36 & (28, 8) & \url{https://www.openml.org/d/1062} \\
		$D_7$ & ILPD & 10 & 583 & (415, 167) & \url{https://www.openml.org/d/1480} \\
		$D_8$ & PLRL & 13 & 315 & (133, 182) & \url{https://www.openml.org/d/915} \\
		$D_9$ & BTum & 9 & 277 & (160, 117) & \url{https://www.openml.org/d/844} \\
		$D_{10}$ & Sleep & 7 & 55 & (29, 26) & \url{https://www.openml.org/d/739} \\
		$D_{11}$ & EMon & 9 & 61 & (29, 32) & \url{https://www.openml.org/d/944} \\
		$D_{12}$ & MC3 & 39 & 161 & (109, 52) & \url{https://www.openml.org/d/1054} \\
		$D_{13}$ & Heart & 13 & 303 & (204 , 99) & \cite{kgl} \\
		$D_{14}$ & Sonar & 60 & 208 & (111, 97) & \url{https://www.openml.org/d/40} \\
		$D_{15}$ & PRel & 12 & 182 & (130, 52) & \url{https://www.openml.org/d/1490} \\
		$D_{16}$ & GDam & 8 & 155 & (106,  49) &\url{https://www.openml.org/d/1026} \\
		$D_{17}$ & CVine & 8 & 52 & (28,  24) & \url{https://www.openml.org/d/815} \\
		\bottomrule
	\end{tabular}
	\label{datasets}
\end{table}

\subsection{Synthetic Data}
\label{simulation}
To assess the performance of the proposed method (ExNRule) under different scenarios, six datasets with binary responses are generated, where each scenario has 5 features and 100 observations. Out of 100 samples, 50 are generated from a distribution with some fix parameter values and they are assigned to a class $0$ and the remaining 50 instances, which are generated from the same distribution with different parameter values, are reserved for class $1$. A detailed description is given in Table \ref{simdata}, where the first column shows the ID of datasets, while the second and third columns represent features' distributions of class $0$ and class $1$, respectively.

\begin{table}[H]
	\centering
	\setlength{\tabcolsep}{2.5pt}
	\fontsize{9.5}{9.5}\selectfont
	\setlength{\tabcolsep}{6pt}
	\renewcommand{\arraystretch}{1.5}
	\caption{Description of the synthetic datasets}
	\begin{tabular}{ccc}
		\toprule
		Scenario ID & Feature's distribution for class $0$ & Feature's distribution for class $1$ \\
		\midrule
		$S_1$ & $Norm(\mu = 5,\sigma = 5)$ & $Norm(\mu = 10,\sigma = 10)$ \\
		$S_2$ & $Norm(\mu = 5,\sigma = 5)$ & $Norm(\mu = 10,\sigma = 5)$ \\
		$S_3$ & $Norm(\mu = 5,\sigma = 5)$ & $Norm(\mu = 10,\sigma = 4)$ \\
		$S_4$ & $Norm(\mu = 5,\sigma = 4)$ & $Norm(\mu = 10,\sigma = 4)$ \\
		$S_5$ & $Norm(\mu = 5,\sigma = 5)$ & $Norm(\mu = 5,\sigma = 10)$ \\
		$S_6$ & $Norm(\mu = 3,\sigma = 3)$ & $Norm(\mu = 1,\sigma = 3)$ \\
		\bottomrule
	\end{tabular}
\label{simdata}
\end{table}

\subsection{Experimental setup}
\label{sec:Experimental setup}
The experimental setup consists of 17 benchmark datasets presented in Section \ref{data5} and 6 synthetic datasets described in Section \ref{simulation}. Each dataset is divided into two mutually exclusive groups, i.e., $70\%$ training and  $30\%$ testing parts. The proposed ExNRule is constructed using 500 individual learners, each on a random bootstrap sample taken from the training observations with a subset of attributes; i.e., $p' = \sqrt{p}$ and $k$ neighbours are selected in an extended manner. The predictions are given in each base model using majority voting. Final prediction is the model value of the results produced by all 500 base learners. The value of $k = 3$ is used to compare the proposed method with the other methods, which include $k$NN, R$k$NN, W$k$NN, RF, OTE and SVM. In addition, the novel ExNRule is also compared for different $k$ values (i.e., $k = 3, 5, 7$) with the different extensions of $k$-nearest neighbour classifier, i.e., $k$NN, R$k$NN and W$k$NN, on five different datasets.

In order to analyse the datasets using the aforementioned methods, various R packages have been utilised. The package \texttt{caret} \cite{caret} implemented in R is used for $k$NN. The R package \texttt{kknn} \cite{kknn} is used for weighted $k$NN, while R library \texttt{rknn} \cite{rknn} is used for random $k$NN.  For random forest, the R library \texttt{randomForest} \cite{randomForest} is used, while for OTE, the R package \texttt{OTE} \cite{OTE} is used. The R library \texttt{kernlab} \cite{kernlab} is used for SVM model. The R function \texttt{tune.knn} is used to fine-tune $k$NN for various values of the hyper-parameter $k$ ,i.e., $k = 1, 2, 3, \ldots, 10$ in the R package \texttt{e1071} \cite{e1071}. Similarly, R$k$NN is fine-tuned by using different values of $k$, i.e., $k = 1, 2, 3, \dots, 10$ and randomly selecting the number of features in ${\sqrt{p}, p/2, p/3, p/4, p/5}$. The remaining setup is kept as given in the R package \texttt{rknn} \cite{rknn}. The R function \texttt{tune.randomForest} in R library \texttt{e1071} \cite{e1071} is used for fine-tuning the hyper-parameters \texttt{nodesize}, \texttt{ntree} and \texttt{mtry}. The same values are used for OTE in package \texttt{OTE} \cite{OTE}. The linear kernel is used for SVM in the R library \texttt{kernlab} \cite{kernlab} with default values of parameters.

\subsection{Results}
\label{sec:Results}
\begin{figure}[h]
	\centering
	\hspace{-1cm}
	\includegraphics[width=1\textwidth, height=1\textheight]{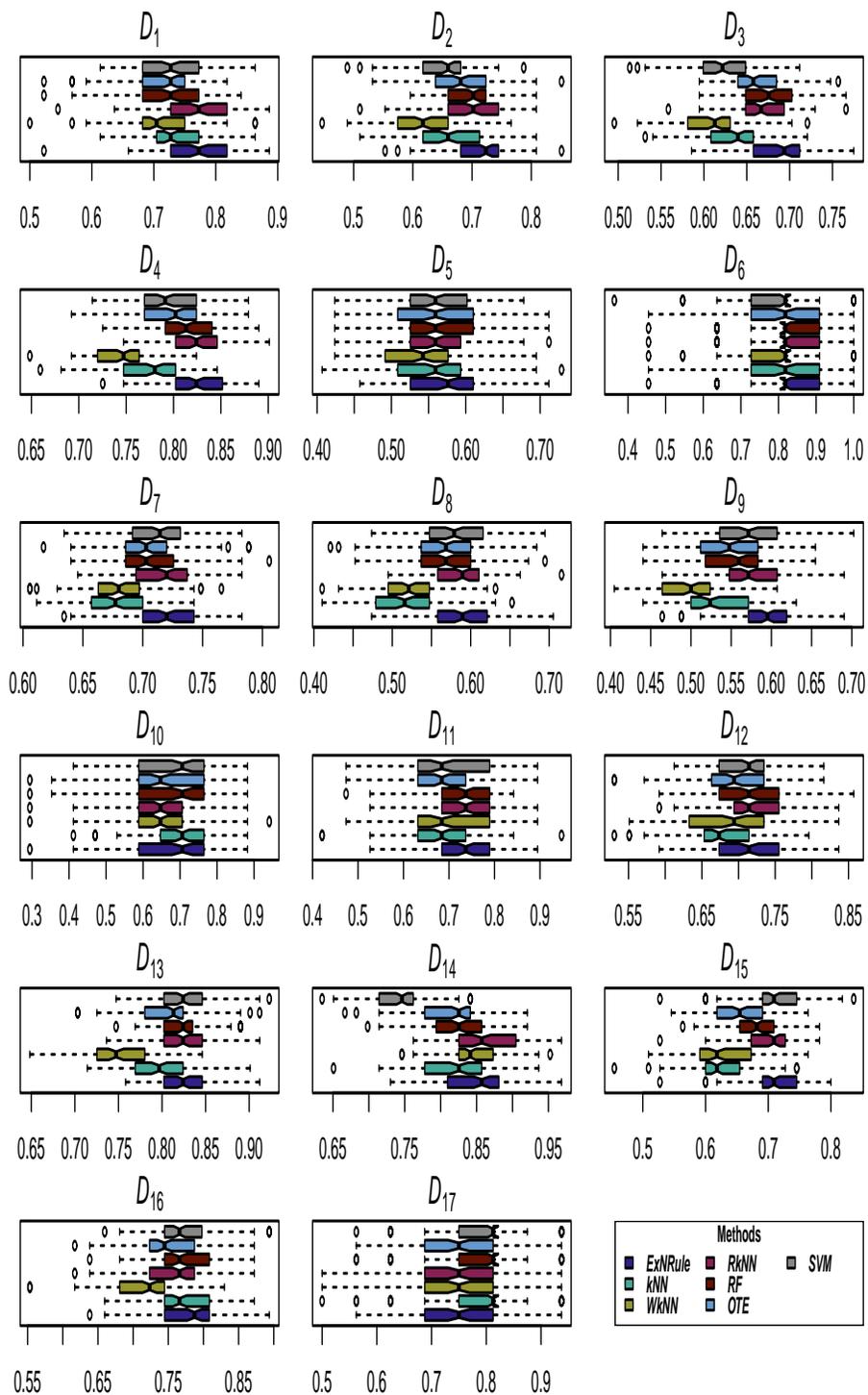}
	\caption{Accuracy of the proposed and other state-of-the-art methods}
	\label{acc1}
\end{figure}

\begin{figure}[h]
	\centering
	\hspace{-1cm}
    \includegraphics[width=1\textwidth, height=1\textheight]{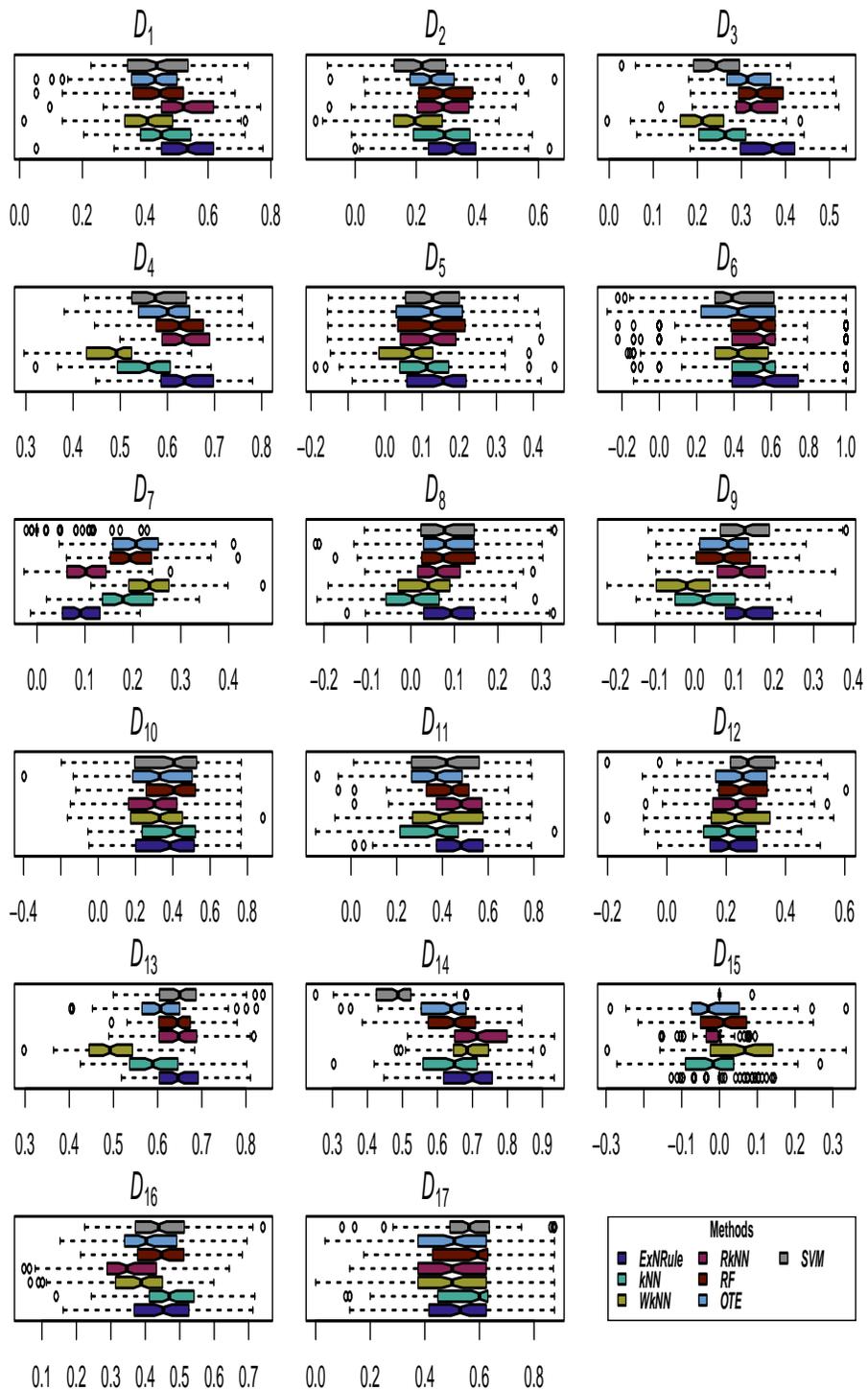}
	\caption{Kappa of the proposed and other state-of-the-art methods}
	\label{kapa1}
\end{figure}

\begin{figure}[h]
	\centering
	\hspace{-1cm}
    \includegraphics[width=1\textwidth, height=1\textheight]{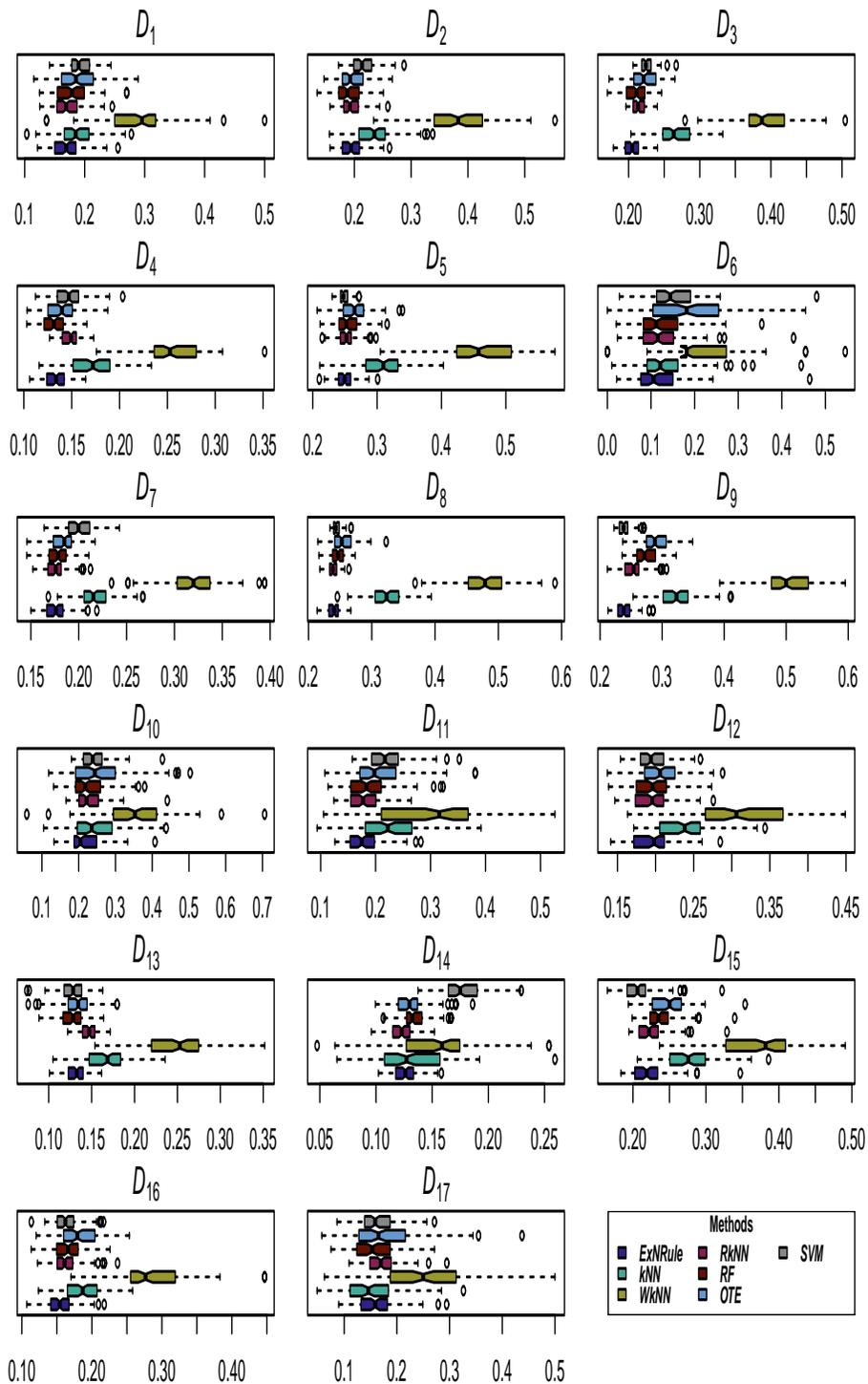}
	\caption{BS of the proposed and other state-of-the-art methods}
	\label{BS1}
\end{figure}

Table \ref{tab1} shows the results given by the proposed ExNRule and the other state-of-the-art methods for 17 datasets. The results reveal that the proposed ExNRule outperforms the rest of procedures on the majority of the datasets. ExNRule gives the highest accuracy as compared to the other procedures on 13 datasets (i.e., $D_1$, $D_2$, $D_3$, $D_4$, $D_5$, $D_6$, $D_7$, $D_8$, $D_9$, $D_{11}$, $D_{12}$, $D_{13}$, $D_{16}$), $k$NN and R$k$NN gives higher accuracy as compared to others on two dataset (i.e., $D_{12}$, $D_{14}$), while W$k$NN yields poor performance. Random forest and OTE do not give optimal results on any of the datasets. SVM performs better than the others on 4 datasets (i.e., $D_{13}$, $D_{15}$, $D_{16}$, $D_{17}$). In terms of Cohen's kappa the proposed method outperforms its competitors on 10 datasets (i.e., $D_1$, $D_2$, $D_3$, $D_4$, $D_5$, $D_6$, $D_8$, $D_9$, $D_{11}$, $D_{13}$). $k$NN and OTE performed poorly on all datasets in terms of kappa. W$k$NN and SVM give higher kappa values on 3  2 datasets, respectively, while R$k$NN outperforms the others on 1 dataset. RF also gives high accuracy on 1 (i.e., $D_10$) dataset. In terms of Brier score ($BS$), ExNRule outperforms the other methods on 7 (i.e. $D_1$, $D_3$, $D_6$, $D_7$, $D_9$, $D_{10}$, $D_{11}$) datasets, while R$k$NN and RF give minimum $BS$ values on 5 and 3 datasets, that is ($D_7$, $D_8$, $D_{11}$, $D_{12}$, $D_{14}$) and ($D_2$, $D_4$, $D_{13}$), respectively. Moreover, SVM outperforms the other methods on 4 datasets (i.e., $D_5$, $D_9$, $D_{15}$, $D_{16}$). Similarly, $k$NN gives a high value on 1 (i.e. $D_{17}$) dataset, while W$k$NN and OTE do not perform better in any of the datasets in terms of $BS$. For further insights into the results, boxplots of the performance metrics are also constructed. Figure \ref{acc1}, \ref{kapa1} and \ref{BS1} show the boxplots of classification accuracy, kappa and Brier score, respectively. The boxtplots also demonstrates that the proposed method is outperforming the others in majority of the cases.

\begin{figure}[h]
	\centering
	\hspace{-1cm}
    \includegraphics[width=1\textwidth, height=1\textheight]{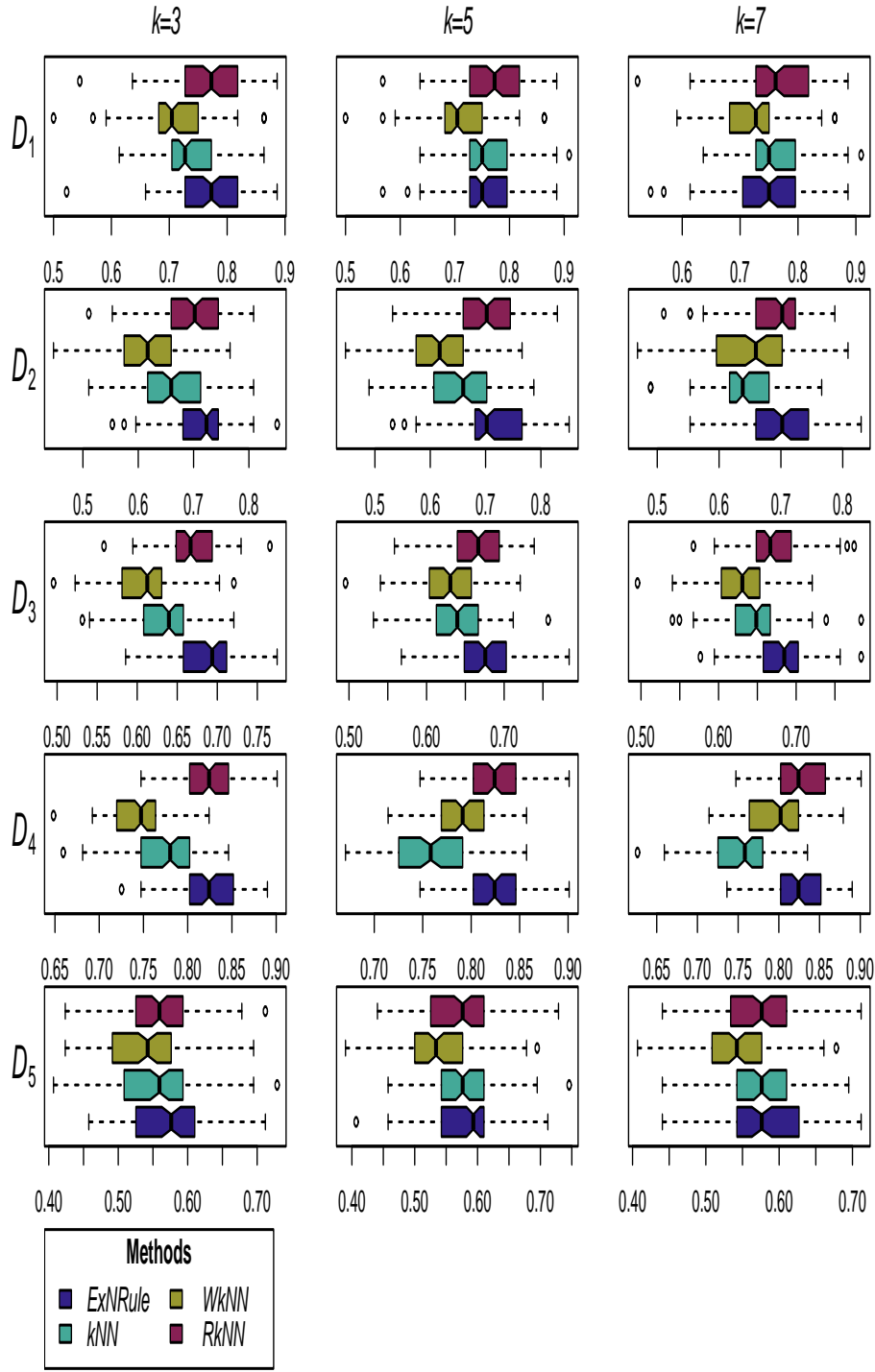}
	\caption{Accuracy of the proposed and other $k$NN based methods for different $k$ values}
	\label{acc2}
\end{figure}

\begin{figure}[h]
	\centering
	\hspace{-1cm}
    \includegraphics[width=1\textwidth, height=1\textheight]{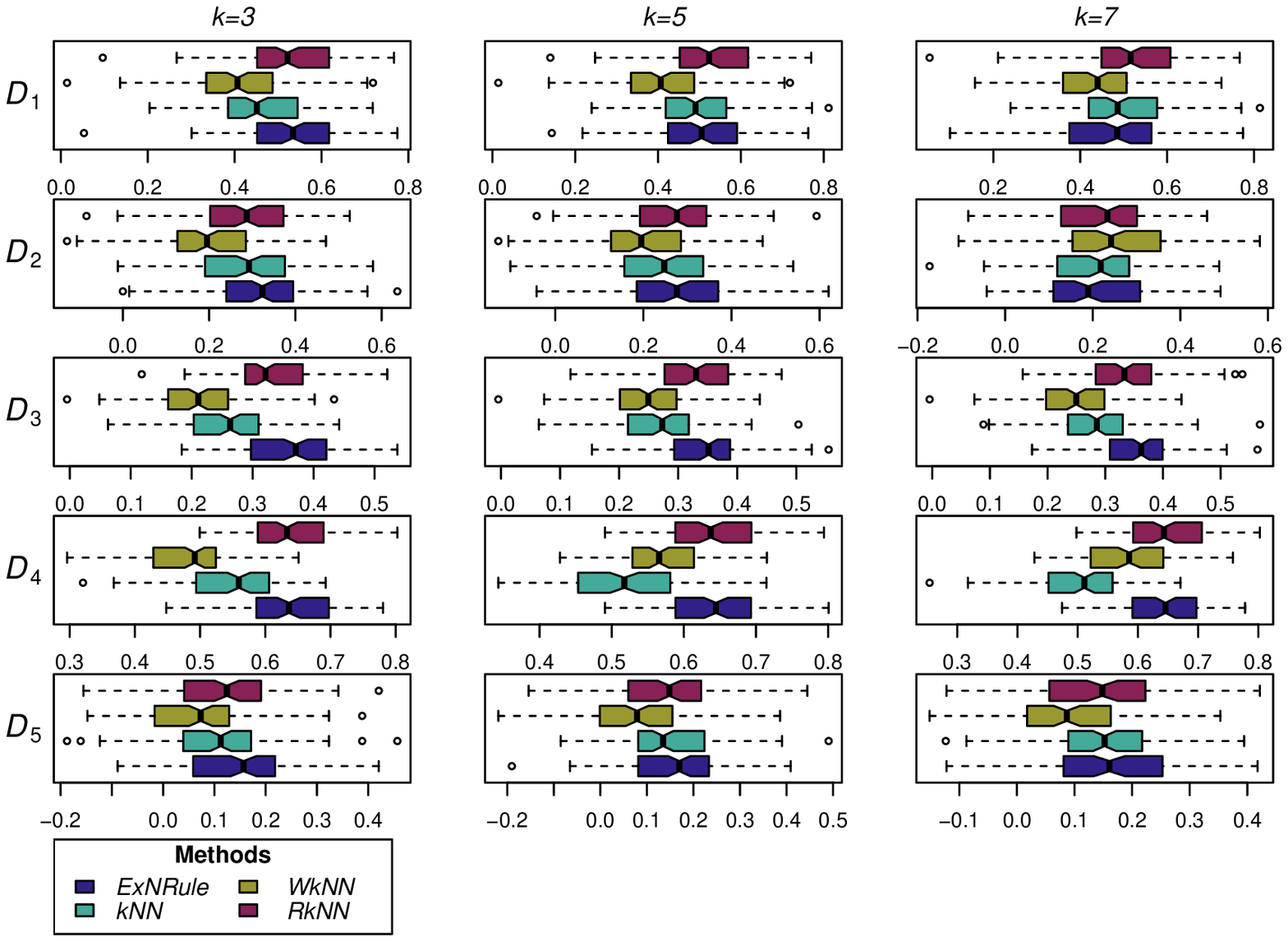}
	\caption{Kappa of the proposed and other $k$NN based methods for different $k$ values}
	\label{kapa2}
\end{figure}

\begin{figure}[h]
	\centering
	\hspace{-1cm}
    \includegraphics[width=1\textwidth, height=1\textheight]{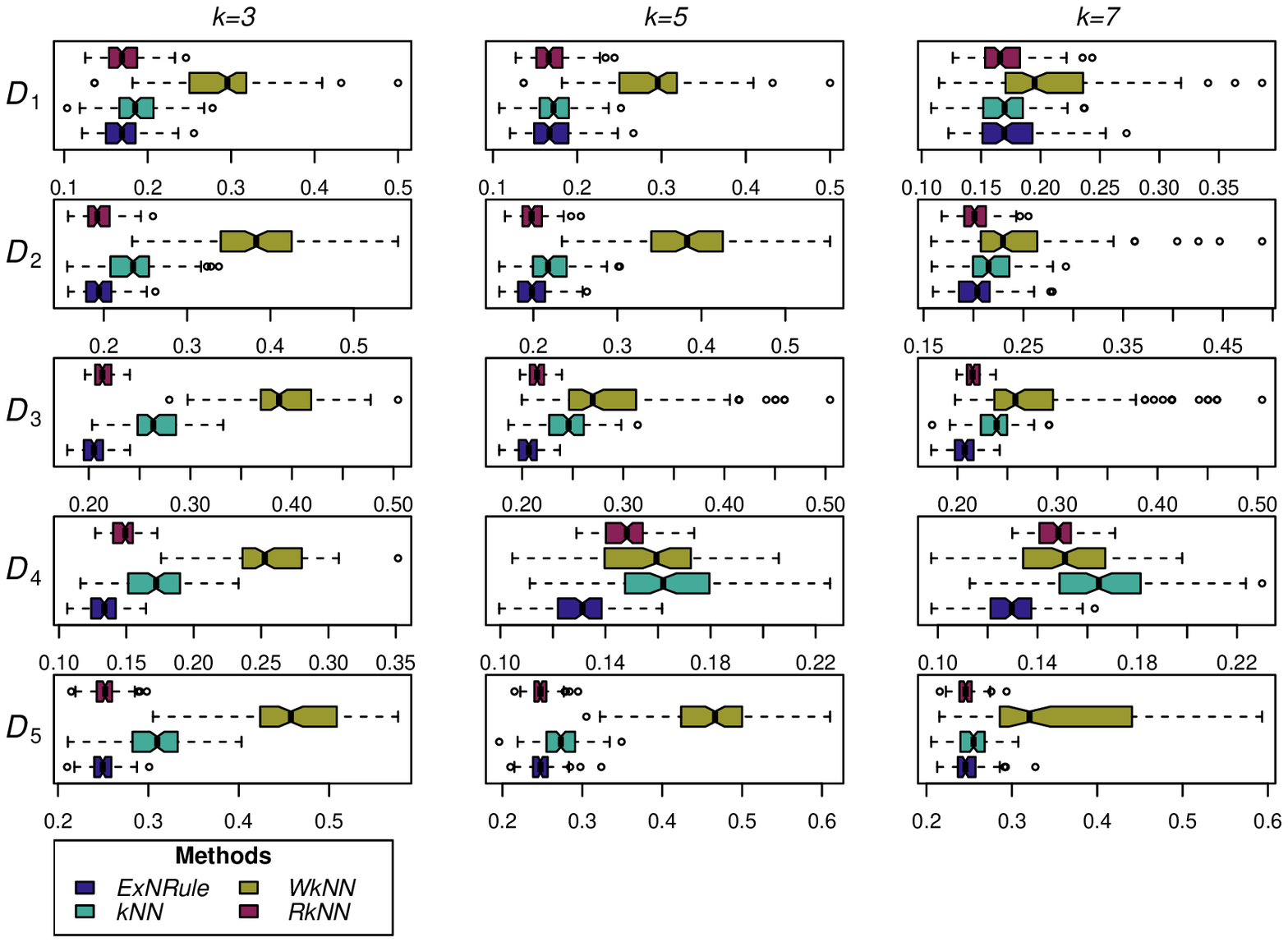}
	\caption{BS of the proposed and other $k$NN based methods for different $k$ values}
	\label{BS2}
\end{figure}

The results of the proposed ExNRule method and other $k$NN based procedures for $k = 3, 5, 7$ are given in Table \ref{tab2} for 5 benchmark datasets. It is clear from the table that the proposed method is not affected by the $k$ parameter as much as the other $k$NN based methods. The ExNRule gives promising results in majority of the datasets in terms of almost all the performance metrics. Boxplots are constructed for accuracy, kappa and BS in Figures \ref{acc2} , \ref{kapa2} and \ref{BS2}, respectively.

\begin{figure}[h]
	\centering
	\hspace{-1cm}
	\includegraphics[width=1\textwidth, height=1\textheight]{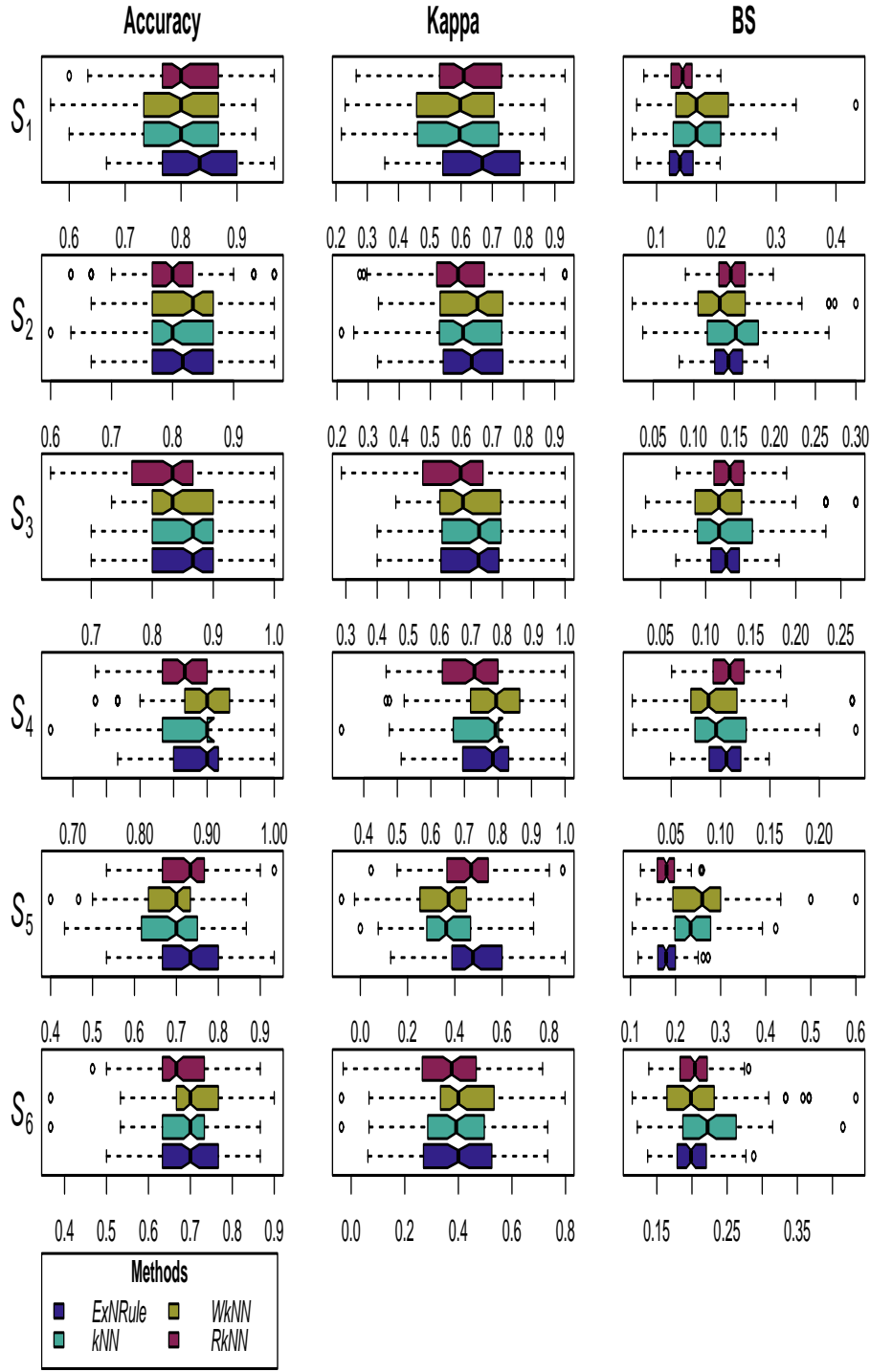}
	\caption{Accuracy, kappa and BS of the proposed and other $k$NN based methods on simulated datasets}
	\label{simplot}
\end{figure}

The results of the ExNRule and other $k$NN based classifiers on synthetic datasets are given in Table \ref{simtab}, which show that the proposed method has outperformed the other competitors in majority of the cases. Particularly, the ExNRule method performs better in a situation where there is more variation in the feature values and where the classes of the observations are not linearly separable. The boxplots for accuracy, kappa and BS are presented in Figure \ref{simplot}. The proposed method did not outperform the other methods in simulation scenarios with small variations in the feature space. This shows that the ExNRule is a recommended method for datasets with diverse patterns.
\begin{table}[!h]
	\centering
	\rotatebox{270}{%
		\begin{minipage}{1\textheight}
			\setlength{\tabcolsep}{1.3pt}
			\fontsize{9.5}{9.5}\selectfont
			\renewcommand{\arraystretch}{1.5}
	\caption{Results of the proposed ExNRule and other state-of-the-art methods on benchmark datasets.}
	
\begin{tabular}{cccccccccccccccccccc}
	\toprule
	\multirow{2}{*}{Metrics} & \multicolumn{1}{c}{\multirow{2}{*}{Methods}} & \multicolumn{17}{c}{Datasets} & \multicolumn{1}{c}{\multirow{2}{*}{Mean}} \\
	\cmidrule(lr){3-19}
	& \multicolumn{1}{c}{} & $D_1$ & $D_2$ & $D_3$ & $D_4$ & $D_5$ & $D_6$ & $D_7$ & $D_8$ & $D_9$ & $D_{10}$ & $D_{11}$ & $D_{12}$ & $D_{13}$ & $D_{14}$ & $D_{15}$ & $D_{16}$ & $D_{17}$ & \multicolumn{1}{c}{} \\
	\midrule
	\multirow{7}{*}{Accuracy} & ExNRule & \textbf{0.768} & \textbf{0.716} & \textbf{0.683} & \textbf{0.824} & \textbf{0.573} & \textbf{0.836} & \textbf{0.719} & \textbf{0.591} & \textbf{0.592} & 0.671 & \textbf{0.733} & \textbf{0.716} & \textbf{0.828} & 0.850 & 0.709 & \textbf{0.778} & 0.769 & \textbf{0.727} \\
	& $k$NN & 0.742 & 0.666 & 0.632 & 0.776 & 0.552 & 0.823 & 0.678 & 0.515 & 0.527 & 0.680 & 0.677 & 0.677 & 0.798 & 0.820 & 0.625 & 0.772 & 0.782 & 0.691 \\
	& W$k$NN & 0.709 & 0.620 & 0.608 & 0.745 & 0.533 & 0.788 & 0.682 & 0.521 & 0.496 & 0.661 & 0.707 & 0.686 & 0.750 & 0.850 & 0.623 & 0.720 & 0.756 & 0.674 \\
	& R$k$NN & 0.764 & 0.699 & 0.669 & 0.823 & 0.562 & 0.830 & 0.717 & 0.583 & 0.577 & 0.643 & 0.732 & \textbf{0.716} & 0.827 & \textbf{0.862} & 0.700 & 0.754 & 0.752 & 0.718 \\
	& RF & 0.723 & 0.696 & 0.676 & 0.814 & 0.568 & 0.825 & 0.707 & 0.570 & 0.550 & \textbf{0.679} & 0.711 & 0.712 & 0.825 & 0.823 & 0.681 & 0.769 & 0.779 & 0.712 \\
	& OTE & 0.716 & 0.678 & 0.664 & 0.801 & 0.563 & 0.790 & 0.703 & 0.567 & 0.544 & 0.652 & 0.693 & 0.701 & 0.808 & 0.814 & 0.653 & 0.749 & 0.763 & 0.698 \\
	& SVM & 0.734 & 0.641 & 0.623 & 0.793 & 0.568 & 0.783 & 0.710 & 0.580 & 0.576 & \textbf{0.679} & 0.698 & 0.706 & \textbf{0.828} & 0.740 & \textbf{0.713} & 0.772 & \textbf{0.786} & 0.702 \\
	
	\multicolumn{1}{l}{} &  &  &  &  &  &  &  &  &  &  &  &  &  &  &  &  &  &  &  \\
	\multirow{7}{*}{Kappa} & ExNRule & \textbf{0.527} & \textbf{0.316} & \textbf{0.360} & \textbf{0.642} & \textbf{0.143} & \textbf{0.542} & 0.092 & \textbf{0.090} & \textbf{0.132} & 0.349 & \textbf{0.467} & 0.222 & \textbf{0.651} & 0.695 & 0.008 & 0.449 & 0.535 & \textbf{0.366} \\
	& $k$NN & 0.465 & 0.283 & 0.253 & 0.550 & 0.105 & 0.517 & 0.186 & 0.001 & 0.026 & 0.363 & 0.355 & 0.202 & 0.591 & 0.634 & -0.026 & \textbf{0.471} & 0.559 & 0.326 \\
	& W$k$NN & 0.406 & 0.201 & 0.211 & 0.483 & 0.065 & 0.430 & \textbf{0.238} & 0.033 & -0.023 & 0.322 & 0.412 & 0.241 & 0.494 & 0.695 & \textbf{0.063} & 0.374 & 0.504 & 0.303 \\
	& R$k$NN & 0.521 & 0.282 & 0.329 & 0.640 & 0.121 & 0.515 & 0.103 & 0.063 & 0.120 & 0.294 & 0.466 & 0.233 & 0.649 & \textbf{0.720} & -0.014 & 0.355 & 0.504 & 0.347 \\
	& RF & 0.440 & 0.295 & 0.344 & 0.623 & 0.131 & 0.497 & 0.195 & 0.081 & 0.073 & \textbf{0.369} & 0.422 & 0.254 & 0.645 & 0.642 & 0.009 & 0.447 & 0.552 & 0.354 \\
	& OTE & 0.429 & 0.255 & 0.318 & 0.597 & 0.121 & 0.392 & 0.207 & 0.081 & 0.075 & 0.315 & 0.386 & 0.253 & 0.610 & 0.623 & -0.014 & 0.410 & 0.518 & 0.328 \\
	& SVM & 0.447 & 0.212 & 0.241 & 0.580 & 0.133 & 0.419 & 0.018 & 0.083 & 0.130 & 0.363 & 0.397 & \textbf{0.280} & 0.650 & 0.477 & 0.001 & 0.447 & \textbf{0.564} & 0.320 \\
	
	\multicolumn{1}{l}{} &  &  &  &  &  &  &  &  &  &  &  &  &  &  &  &  &  &  & \textbf{} \\
	\multirow{7}{*}{BS} & ExNRule & \textbf{0.170} & 0.195 & \textbf{0.205} & 0.134 & 0.251 & \textbf{0.116} & \textbf{0.176} & 0.240 & \textbf{0.239} & \textbf{0.218} & \textbf{0.179} & 0.196 & 0.132 & 0.126 & 0.222 & \textbf{0.156} & 0.162 & \textbf{0.183} \\
	& $k$NN & 0.186 & 0.236 & 0.266 & 0.172 & 0.310 & 0.132 & 0.218 & 0.324 & 0.324 & 0.243 & 0.227 & 0.237 & 0.167 & 0.131 & 0.278 & 0.186 & \textbf{0.151} & 0.223 \\
	& W$k$NN & 0.291 & 0.380 & 0.392 & 0.255 & 0.467 & 0.212 & 0.318 & 0.479 & 0.504 & 0.339 & 0.293 & 0.314 & 0.250 & 0.150 & 0.377 & 0.280 & 0.244 & 0.326 \\
	& R$k$NN & 0.172 & 0.195 & 0.215 & 0.148 & 0.253 & 0.121 & \textbf{0.176} & \textbf{0.239} & 0.254 & 0.229 & \textbf{0.179} & \textbf{0.195} & 0.147 & \textbf{0.123} & 0.225 & 0.164 & 0.172 & 0.189 \\
	& RF & 0.179 & \textbf{0.189} & 0.211 & \textbf{0.132} & 0.255 & 0.124 & 0.178 & 0.247 & 0.274 & 0.227 & 0.186 & 0.196 & \textbf{0.127} & 0.135 & 0.238 & 0.166 & 0.157 & 0.189 \\
	& OTE & 0.187 & 0.197 & 0.222 & 0.138 & 0.264 & 0.184 & 0.183 & 0.255 & 0.292 & 0.258 & 0.206 & 0.206 & 0.133 & 0.131 & 0.248 & 0.181 & 0.178 & 0.204 \\
	& SVM & 0.193 & 0.217 & 0.226 & 0.148 & \textbf{0.249} & 0.153 & 0.200 & 0.245 & \textbf{0.239} & 0.242 & 0.221 & 0.197 & 0.128 & 0.178 & \textbf{0.208} & 0.165 & 0.165 & 0.198\\
	\bottomrule
\end{tabular}

\label{tab1}
\end{minipage}}
\end{table}
\twocolumn
\onecolumn

\begin{table}[h]
	\centering
	\rotatebox{270}{%
		\begin{minipage}{1\textheight}
			\hspace{-1cm}
			\vspace{-1cm}
			\setlength{\tabcolsep}{2pt}
			\fontsize{9.5}{9.5}\selectfont
			\renewcommand{\arraystretch}{1.5}
			\caption{Results of the proposed ExNRule and $k$NN based methods for different values of $k$.}
\begin{tabular}{cccccccccccccccccc}
		 	\toprule
		 	\multirow{3}{*}{Metrics} & \multicolumn{1}{c}{\multirow{3}{*}{Methods}} & \multicolumn{15}{c}{Datasets} & \multicolumn{1}{c}{\multirow{3}{*}{Mean}} \\
		 	\cline{3-17}
		 	& \multicolumn{1}{c}{} & \multicolumn{3}{c}{$D_1$} & \multicolumn{3}{c}{$D_2$} & \multicolumn{3}{c}{$D_3$} & \multicolumn{3}{c}{$D_4$} & \multicolumn{3}{c}{$D_5$} & \multicolumn{1}{c}{} \\ \cmidrule(lr){3-5}\cmidrule(lr){9-11} \cmidrule(lr){15-17} \cmidrule(lr){6-8} \cmidrule(lr){12-14}
		 	& \multicolumn{1}{c}{} & $k = 3$ & $k = 5$ & $k = 7$ & $k = 3$ & $k = 5$ & $k = 7$ & $k = 3$ & $k = 5$ & $k = 7$ & $k = 3$ & $k = 5$ & $k = 7$ & $k = 3$ & $k = 5$ & $k = 7$ & \multicolumn{1}{c}{} \\ \midrule
		 	\multirow{4}{*}{Accuracy} & EXNRule & \textbf{0.768} & 0.759 & 0.745 & \textbf{0.716} & \textbf{0.709} & \textbf{0.694} & \textbf{0.683} & \textbf{0.677} & \textbf{0.681} & \textbf{0.824} & \textbf{0.825} & 0.825 & \textbf{0.573} & \textbf{0.584} & \textbf{0.588} & \textbf{0.710} \\
		 	& $k$NN & 0.742 & 0.757 & 0.758 & 0.666 & 0.657 & 0.650 & 0.632 & 0.641 & 0.646 & 0.776 & 0.761 & 0.752 & 0.552 & 0.575 & 0.573 & 0.676 \\
		 	& W$k$NN & 0.709 & 0.709 & 0.727 & 0.620 & 0.620 & 0.649 & 0.608 & 0.629 & 0.630 & 0.745 & 0.787 & 0.794 & 0.533 & 0.538 & 0.546 & 0.656 \\
		 	& R$k$NN & 0.764 & \textbf{0.766} & \textbf{0.764} & 0.699 & 0.700 & 0.690 & 0.669 & 0.669 & 0.671 & 0.823 & 0.824 & \textbf{0.827} & 0.562 & 0.571 & 0.576 & 0.705 \\
		 	\multicolumn{1}{l}{} &  &  &  &  &  &  &  &  &  &  &  &  &  &  &  &  &  \\
		 	\multirow{4}{*}{Kappa} & EXNRule & \textbf{0.527} & 0.506 & 0.474 & \textbf{0.316} & \textbf{0.269} & 0.208 & \textbf{0.360} & \textbf{0.347} & \textbf{0.357} & \textbf{0.642} & \textbf{0.644} & 0.644 & \textbf{0.143} & \textbf{0.165} & \textbf{0.174} & \textbf{0.385} \\
		 	& $k$NN & 0.465 & 0.498 & 0.501 & 0.283 & 0.245 & 0.205 & 0.253 & 0.273 & 0.287 & 0.550 & 0.520 & 0.500 & 0.105 & 0.151 & 0.148 & 0.332 \\
		 	& W$k$NN & 0.406 & 0.406 & 0.441 & 0.201 & 0.201 & \textbf{0.242} & 0.211 & 0.252 & 0.253 & 0.483 & 0.571 & 0.584 & 0.065 & 0.076 & 0.094 & 0.299 \\
		 	& R$k$NN & 0.521 & \textbf{0.525} & \textbf{0.522} & 0.282 & 0.265 & 0.219 & 0.329 & 0.331 & 0.336 & 0.640 & 0.643 & \textbf{0.647} & 0.121 & 0.137 & 0.146 & 0.378 \\
		 	\multicolumn{1}{l}{} &  &  &  &  &  &  &  &  &  &  &  &  &  &  &  &  &  \\
		 	\multirow{4}{*}{BS} & EXNRule & \textbf{0.170} & 0.171 & 0.173 & \textbf{0.195} & 0.200 & 0.204 & \textbf{0.205} & \textbf{0.206} & \textbf{0.207} & \textbf{0.134} & \textbf{0.131} & \textbf{0.130} & \textbf{0.251} & 0.250 & 0.249 & \textbf{0.192} \\
		 	& $k$NN & 0.186 & 0.171 & \textbf{0.169} & 0.236 & 0.221 & 0.219 & 0.266 & 0.247 & 0.238 & 0.172 & 0.164 & 0.165 & 0.310 & 0.273 & 0.256 & 0.220 \\
		 	& W$k$NN & 0.291 & 0.291 & 0.207 & 0.380 & 0.380 & 0.244 & 0.392 & 0.292 & 0.283 & 0.255 & 0.157 & 0.150 & 0.467 & 0.462 & 0.359 & 0.307 \\
		 	& R$k$NN & 0.172 & \textbf{0.169} & \textbf{0.169} & \textbf{0.195} & \textbf{0.199} & \textbf{0.202} & 0.215 & 0.215 & 0.216 & 0.148 & 0.148 & 0.148 & 0.253 & \textbf{0.249} & \textbf{0.247} & 0.196\\
		 	\bottomrule
		 \end{tabular}
\label{tab2}
\end{minipage}}
\end{table}

\begin{table}[h]
	\vspace{-6cm}
	\centering
	\caption{Comparison of the proposed ExNRule with the other classical $k$NN and its derivatives based on synthetic datasets}
	\begin{tabular}{cccccccc}
		\toprule
		\multirow{2}{*}{Metrics} & \multicolumn{1}{c}{\multirow{2}{*}{Methods}} & \multicolumn{6}{c}{Senarios} \\
		\cmidrule(lr){3-8}
		& \multicolumn{1}{c}{} & $S_1$ & $S_2$ & $S_3$ & $S_4$ & $S_5$ & $S_6$ \\
		\midrule
		\multirow{4}{*}{Accuracy} & ExNRule & \textbf{0.832} & \textbf{0.823} & \textbf{0.852} & 0.884 & \textbf{0.742} & 0.693 \\
		& $k$NN & 0.786 & 0.811 & 0.850 & 0.878 & 0.682 & 0.696 \\
		& W$k$NN & 0.789 & 0.821 & 0.849 & \textbf{0.887} & 0.680 & \textbf{0.706} \\
		& R$k$NN & 0.809 & 0.798 & 0.833 & 0.862 & 0.730 & 0.675 \\
		\multicolumn{1}{l}{} &  &  &  &  &  &  &  \\
		\multirow{4}{*}{Kappa} & ExNRule & \textbf{0.666} & \textbf{0.644} & \textbf{0.702} & 0.766 & \textbf{0.493} & 0.396 \\
		& $k$NN & 0.574 & 0.619 & 0.696 & 0.752 & 0.372 & 0.393 \\
		& W$k$NN & 0.581 & 0.640 & 0.695 & \textbf{0.772} & 0.363 & \textbf{0.412} \\
		& R$k$NN & 0.618 & 0.594 & 0.664 & 0.722 & 0.465 & 0.358 \\
		\multicolumn{1}{l}{} &  &  &  &  &  &  &  \\
		\multirow{4}{*}{BS} & ExNRule & \textbf{0.141} & 0.142 & 0.122 & 0.104 & 0.183 & \textbf{0.200} \\
		& $k$NN & 0.169 & 0.149 & 0.121 & 0.099 & 0.239 & 0.223 \\
		& W$k$NN & 0.179 & \textbf{0.136} & \textbf{0.120} & \textbf{0.093} & 0.262 & 0.208 \\
		& R$k$NN & 0.142 & 0.147 & 0.126 & 0.107 & \textbf{0.180} & 0.204\\
		\bottomrule
	\end{tabular}
\label{simtab}
\end{table}



\twocolumn
\onecolumn
\section{Conclusion}
\label{conclusion}
This paper presented a $k$ nearest neighbour based ensemble where the neighbours are determined in $k$ steps. Starting from the first nearest observation of the test point, the algorithm identifies a single observation that is closest to the observation at the previous step. At each base model in the ensemble, this search is extended to k steps based on a bootstrap sample with randomly selected subset of the given features. The final predicted class of the test point is determined by using majority vote in the predicted classes given by all the base models. The proposed ensemble is compared with base $k$NN, weighted $k$NN, random $k$NN, random forest, optimal trees ensemble and support vector machine on 17 datasets. Classification accuracy, Cohen's kappa and Brier score are used as performance measures. It has been observed from the results of the analyses that the proposed method, the ExNRule, outperformed the other procedures in the majority of the cases.

The main intuition behind the prediction accuracy of the proposed method is the selection of nearest neighbours in a stepwise pattern. Models based on the ordinary $k$NN might not work well in situations when the test observation follows the pattern of data points with the same class that lie on a certain path not contained in the given sphere. The proposed ensemble fixes this problem. Moreover, the ordinary $k$NN based models are affected by the hyper-parameter $k$, while the proposed method is robust to the choice of $k$. It is shown that for $k = 3, 5, 7$, that neighbours selection of a test point are not affecting performance of the proposed method in majority of the cases. Moreover, the performance of the novel method is also assessed through simulated data and gives optimal results in majority of the cases.

Furthermore, each base learner in the proposed ensemble constructed on a random bootstrap sample drawn from training observations with a randomly selected subset of attributes ensure diversity in the model. 

The proposed method consists of a large number of base models i.e. $B$ and fits $k$NN repeatedly, hence it is time consuming and laborious as compared to ordinary $k$NN. To overcome this issue, one possibility is to parallelize Steps 8-14 of Algorithm 1, for instance, using the \texttt{parallel} \cite{parallel} R package. Performance of the proposed method could further be improved by using appropriate distance formula to determine the paths. Another possible way to improve performance of the method is to use the feature selection procedures as given in \cite{sun2020prediction, hu2020machinery, khan2019feature, chatterjee2020late, hamraz2021robust, mishra2011robust, li2016selection}. This could be used for selecting a set of features from the total feature space for model construction.


\bibliography{mybibfile}
\end{document}